%% file: ms.tex
\documentclass[10pt,twocolumn,letterpaper]{article}
\pdfoutput=1 
\usepackage{cvpr}
\usepackage{times}
\usepackage{epsfig}
\usepackage{graphicx}
\usepackage{amsmath}
\usepackage{amssymb}

\usepackage{xcolor}
\usepackage{bm}
\usepackage{booktabs}
\usepackage{adjustbox}
\usepackage{authblk}
\usepackage[numbers,sort&compress]{natbib}
\usepackage[page]{appendix}
\usepackage[resetlabels]{multibib}
\newcites{supplement}{References (Supplementary Material)}

\cvprfinalcopy % *** Uncomment this line for the final submission

\def\cvprPaperID{10016} % *** Enter the CVPR Paper ID here

\newcommand{\sect}[1]{Section \ref{#1}}

\newcommand{\ten}[1]{\bm{\mathsf{#1}}}
\newcommand{\mat}[1]{\mathbf{#1}}

\newcommand{\base}{Paix\~ao-b\xspace}
\newcommand{\baseposs}{Paix\~ao-b's\xspace}

\begin{document}

%%%%%%%%% TITLE
\title{Fast(er) Reconstruction of Shredded Text Documents\\via Self-Supervised Deep Asymmetric Metric Learning}

\author[1,2]{Thiago M. Paix\~ao\thanks{Corresponding author: paixao@gmail.com.}}
\author[2]{Rodrigo F. Berriel}
\author[2]{Maria C. S. Boeres}
\author[3]{Alessandro L. Koerich}
\author[2]{\\Claudine Badue}
\author[2]{Alberto F. De Souza}
\author[2]{Thiago Oliveira-Santos}
\affil[1]{{\small Federal Institute of Esp\'irito Santo (IFES), Serra, Brazil}}
\affil[2]{{\small Federal University of Esp\'irito Santo (UFES), Vit\'oria, Brazil}}
\affil[3]{{\small \'Ecole de Technologie Sup\'erieure (\'ETS), Montreal, Canada}}
% \affil[ ]{{\tt\small paixao@gmail.com}, {\tt\small berriel@lcad.inf.ufes.br}, {\tt\small boeres@inf.ufes.br}, {\tt\small alessandro.koerich@etsmtl.ca}, {\tt\small \{claudine,alberto,todsantos\}@lcad.inf.ufes.br}}

\renewcommand\Authands{ and }

\maketitle
%\thispagestyle{empty}

%%%%%%%%% ABSTRACT
\begin{abstract}
   The reconstruction of shredded documents consists in arranging the pieces of paper (shreds) in order to reassemble the original aspect of such documents. This task is particularly relevant for supporting forensic investigation as documents may contain criminal evidence. As an alternative to the laborious and time-consuming manual process, several researchers have been investigating ways to perform automatic digital reconstruction. A central problem in automatic reconstruction of shredded documents is the pairwise compatibility evaluation of the shreds, notably for %``black-and-white''
   binary text documents. In this context, deep learning has enabled great progress for accurate reconstructions in the domain of mechanically-shredded documents. A sensitive issue, however, is that current deep model solutions require an inference whenever a pair of shreds has to be evaluated. This work proposes a scalable deep learning approach for measuring pairwise compatibility in which the number of inferences scales linearly (rather than quadratically) with the number of shreds. Instead of predicting compatibility directly, deep models are leveraged to asymmetrically project the raw shred content onto a common metric space in which distance is proportional to the compatibility. Experimental results show that our method has accuracy comparable to the state-of-the-art with a speed-up of about $22$ times for a test instance with $505$ shreds ($20$ mixed shredded-pages from different documents).
\end{abstract}

\input{secs/1_introduction}
\input{secs/2_problem}
\input{secs/3_proposed}
\input{secs/4_experimental}
\input{secs/5_results}
\input{secs/6_conclusion}
\input{secs/acknowledgements}
\nocite{*}
{\small
\bibliographystyle{ieee_fullname}
\bibliography{ms}
}

\onecolumn
\begin{appendices}
\section{Local Samples Nearest Neighbors}

\begin{figure}[h]
	\centering
	\includegraphics[width=0.6\textwidth]{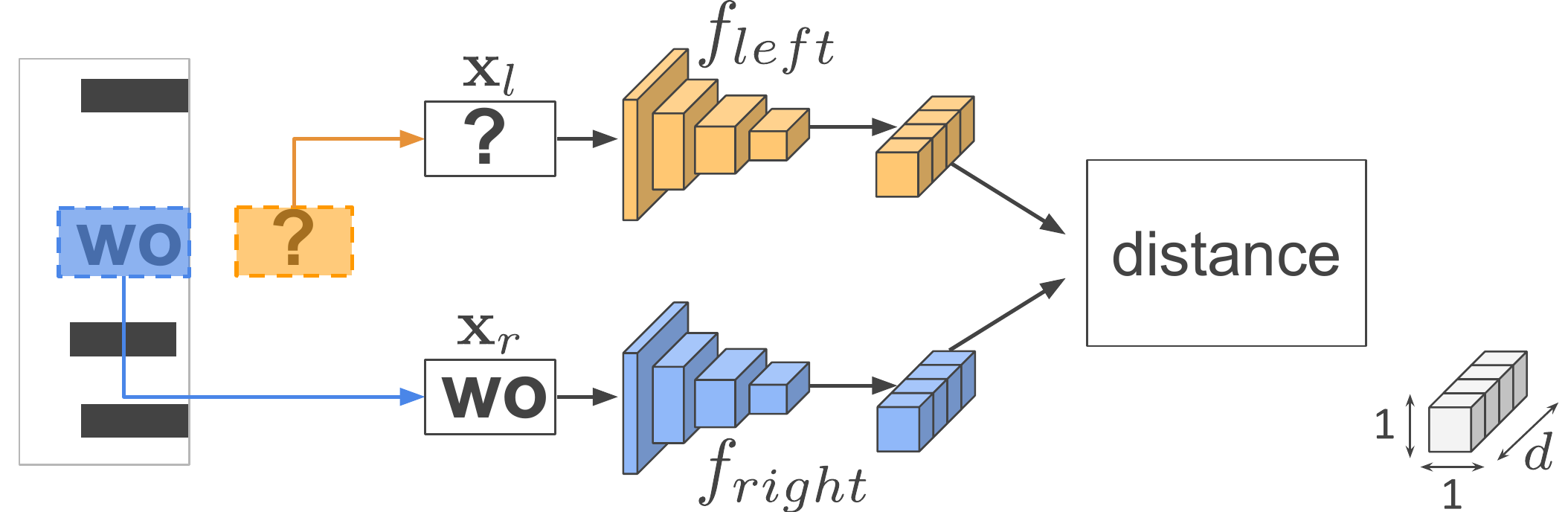}
	\caption{Querying $\mat{x}_l$ samples by fixing $\mat{x}_r$.} 
	\label{fig:query}
\end{figure}

An interesting way to verify how the models are pairing complementary patterns is by fixing $32 \times 32$ samples (query samples) from one of the boundaries and recovering samples of the complementary side. As illustrated in Figure \ref{fig:query}, one can select $\mat{x}_r$ as query sample and try to recover the top-$1$ $\mat{x}_l$'s, i.e., that sample of the left boundary which minimizes the distance to the anchor in the embedding space. Figure 9 in our manuscript has shown some queries for both $\mat{x}_r$ and $\mat{x}_l$ restricted to one shredded document of the test collection. Here, we mixed samples from 3 documents and, similarly, show 28 query samples and their respective top-$3$ complementary samples (distance increasing from top to bottom).

\begin{figure}[h]
	\centering
	\includegraphics[width=\textwidth]{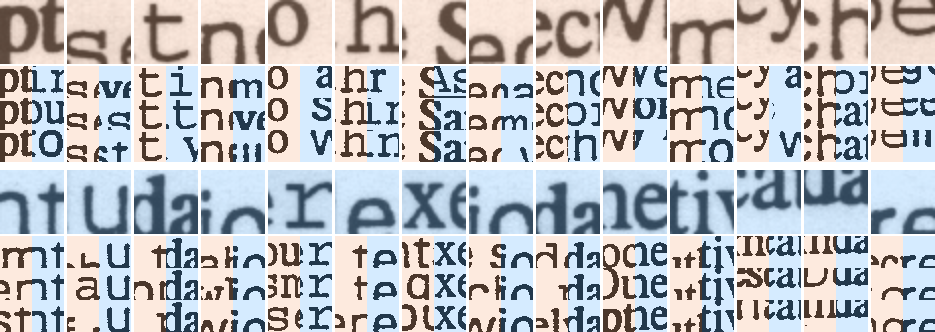}
	\caption{Local samples nearest neighbors. In the top row, the largest square is the ``query'' sample (before binarization) followed, below, by its binary version and its 3 nearest neighbors side-by-side (with the closest in the top row). The blue and orange samples were projected by $f_{right}$ and $f_{left}$, respectively. The bottom row shows some examples in which the ``query'' is projected by the $f_{left}$ instead.}
	\label{fig:query_results}
\end{figure}

\pagebreak

\section{Reconstruction of D2}

\begin{figure}[h]
	\centering
	\includegraphics[width=0.95\textwidth]{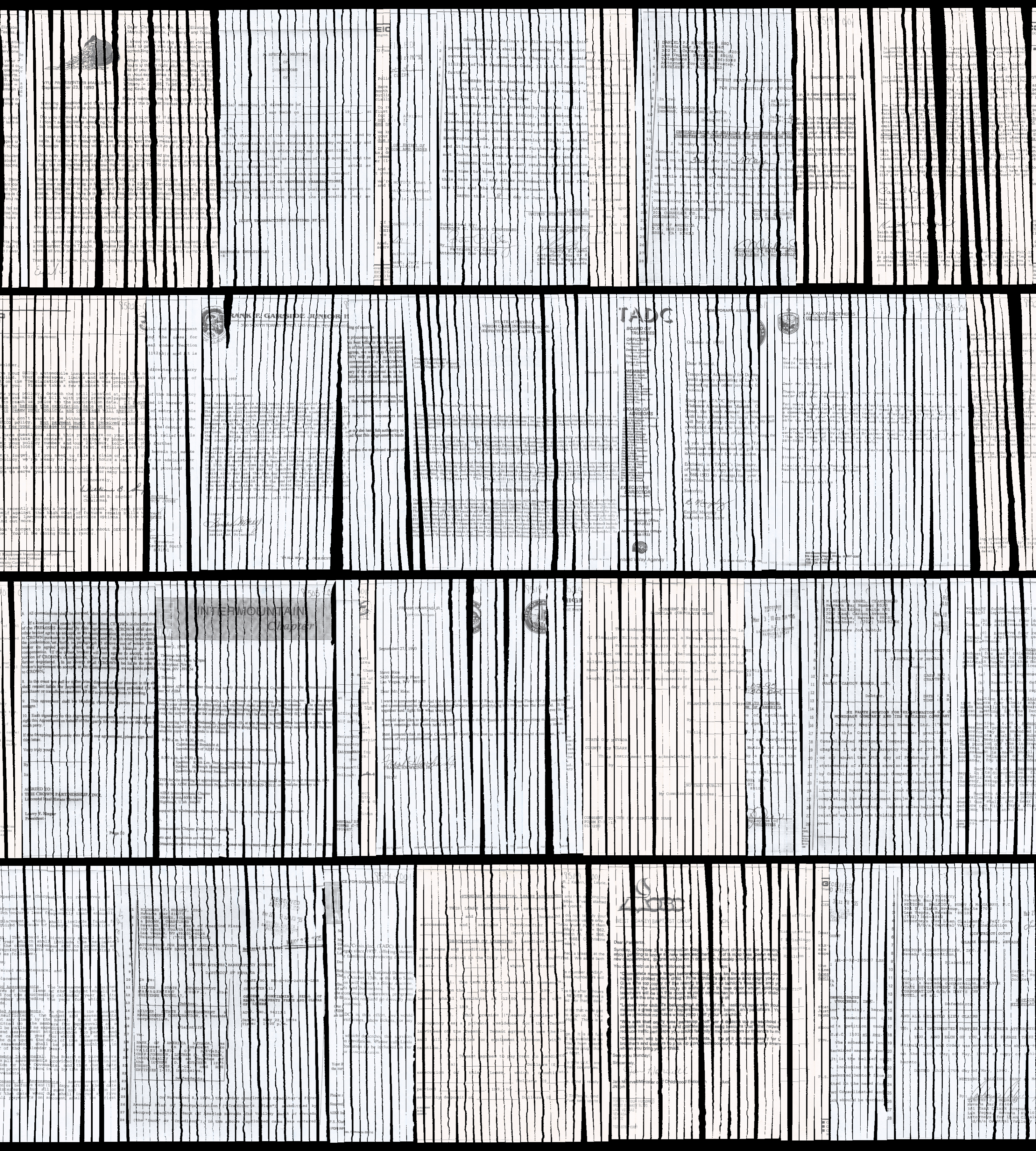}
	\caption{Reconstruction of D2. The generated image was split into 4 parts for better visualization.} 
	\label{fig:reconstruction}
\end{figure}

The dataset D2 comprises $20$ single-page documents, totaling $505$ shreds. 
Figure \ref{fig:reconstruction} shows the reconstruction of the entire D2 dataset, i.e., after mixing all shreds. The shreds were placed side-by-side according to the solution (permutation) computed with the proposed metric learning-based method which achieved the accuracy of $97.22\%$. The pairwise compatibility evaluation took less than $4$ minutes.

\section{Embedding Space}

Figure \ref{fig:embedding} of our manuscript illustrates the embedding space onto which the local samples are projected. For a more concrete view of this space, four charts (Figures \ref{fig:case1} to \ref{fig:case4}) were plotted showing local embeddings produced from a real-shredded document (25 shreds). For each chart, there is a single anchor embedding (in blue), which was produced from an anchor sample $\mat{x}_r$ randomly cropped from the right boundary of an arbitrary shred. The other points (embeddings) in the chart (in orange) corresponds to the samples from the other 24 shreds vertically aligned with the anchor sample, i.e., those which are candidates to match the anchor sample. Notice that the embeddings are numbered according to the shred they belong to, being $0, 1, 2, \ldots, 24$ the ground-truth order of the document. Therefore, the anchor (blue point) indicated by $s$ should match the embedding (orange point) indicated by $s+1$ (a dashed line linking the respective points was made in each chart). For 2-D visualization, embeddings in the original space ($\mathbb{R}^{128}$) were projected to the plane by using T-SNE \citesupplement{maaten2008visualizing_app,van2014accelerating_app}. It is worthy to mention that we analyzed the produced charts to ensure that pairwise distances in $\mathbb{R}^{2}$ are roughly consistent with those in the original space. Also, no vertical alignment between shreds was performed.

\subsection{Case 1}
\begin{figure}[h]
	\centering
	\includegraphics[width=0.6\textwidth]{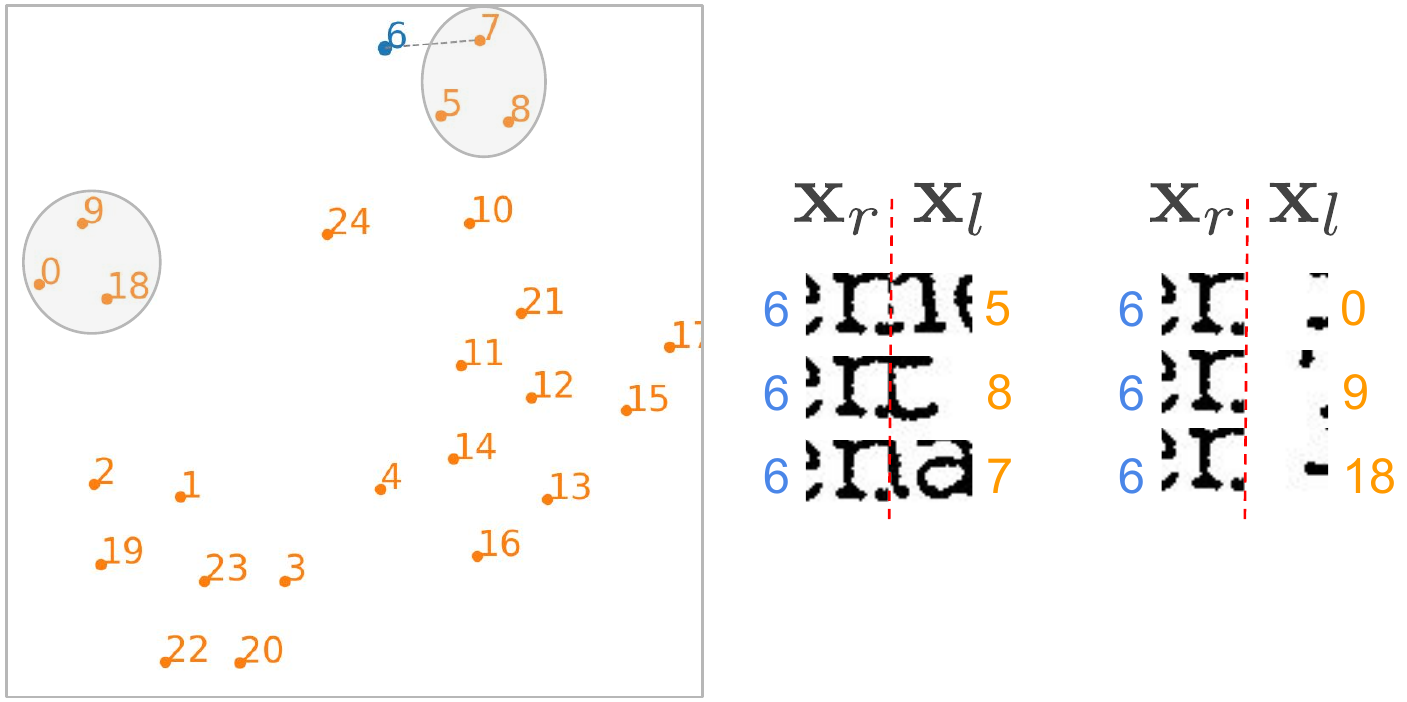}
	\caption{Case 1.} 
	\label{fig:case1}
\end{figure}

In Figure \ref{fig:case1}, samples from two clusters ($\{5, 7, 8\}$ and $\{0, 9, 18\}$) were shown at the right side of the 2-D chart. Although the pairing $(6, 8)$ looks incompatible based on the knowledge of the Latin alphabet, we noticed that the vertical alignment and the emerging horizontal were essential for their close positioning. For the cluster $\{0, 9, 18\}$, it is interesting to note that the information (black pixels) in the $\mat{x}_l$ samples is concentrated in the last columns.

\subsection{Case 2}
\begin{figure}[htb]
	\centering
	\includegraphics[width=0.6\textwidth]{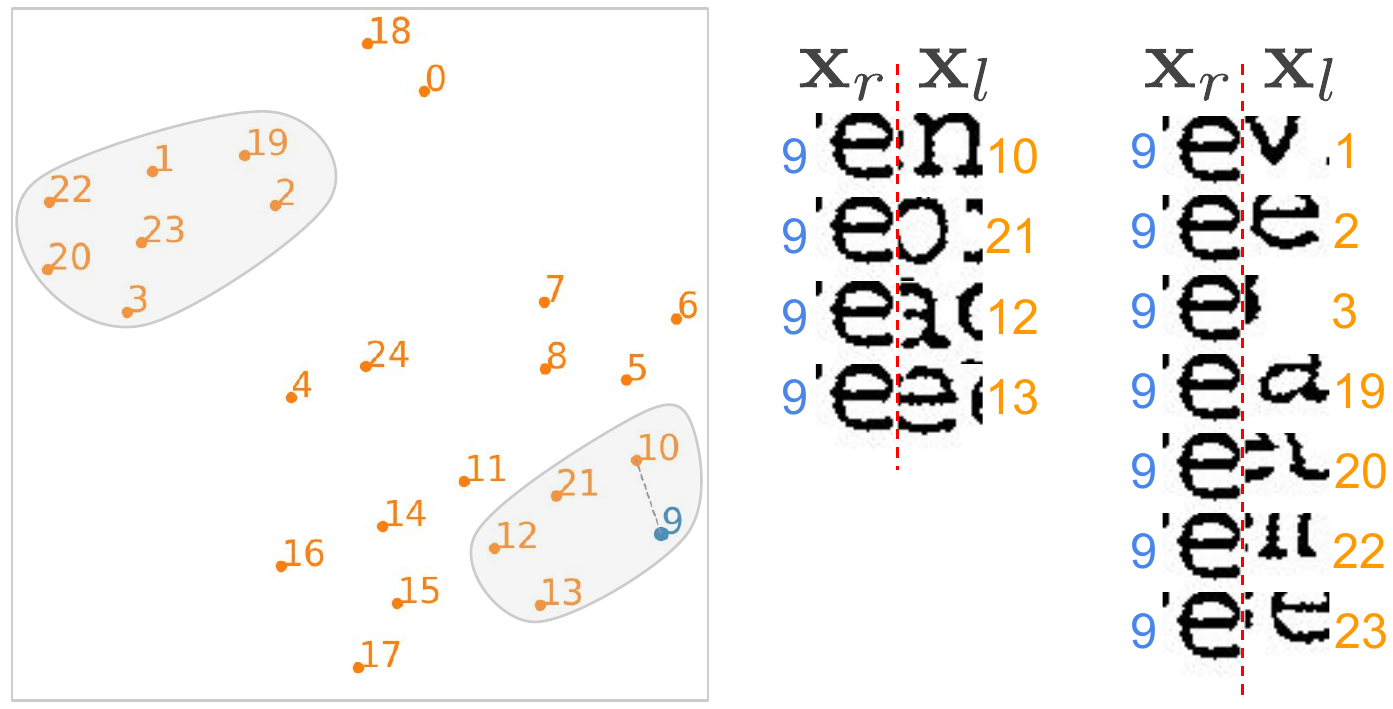}
	\caption{Case 2.} 
	\label{fig:case2}
\end{figure}

% Baseline => https://en.wikipedia.org/wiki/Baseline_(typography)
In Figure \ref{fig:case2}, two clusters were illustrated. As in the previous case, the vertical alignment plays an import role in the positioning of the embeddings. From the cluster $\{1, 2, 3, 19, 20, 22, 23\}$, it can be observed that the $\mat{x}_l$ samples are similarly shifted up compared to the baseline of the anchor. Finally, although the unrealistic pairing $(9, 12)$ yields a distance superior to $(9, 10)$, they are kept close due to the vertical alignment and the emerging connections (three horizontal lines).

\subsection{Case 3}
\begin{figure}[htb]
	\centering
	\includegraphics[width=0.6\textwidth]{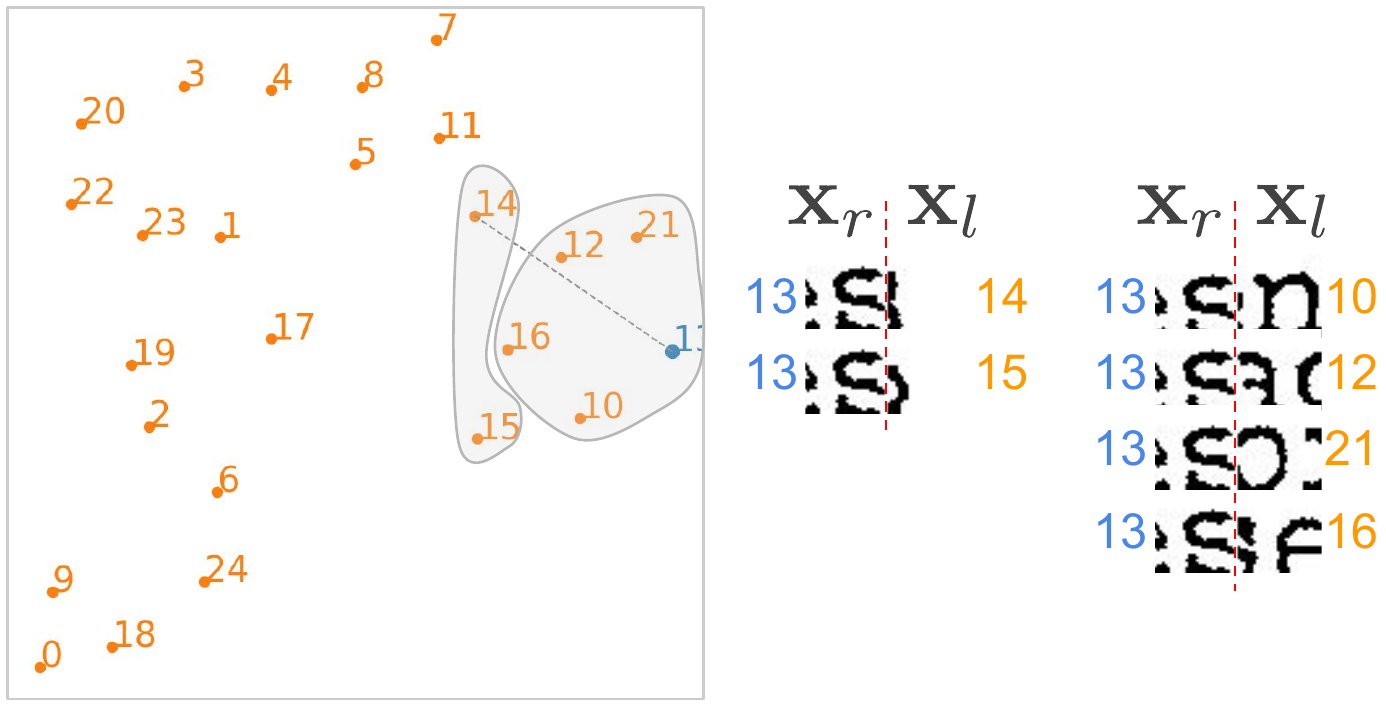}
	\caption{Case 3.} 
	\label{fig:case3}
\end{figure}

The third case, illustrated in Figure \ref{fig:case3}, depicts a situation where a couple of matchings are better evaluated than the corrected one, i.e., $(13, 14)$. In addition to the realistic appearance of the competitors (pairings formed with samples in $\{10, 12, 16, 21\}$), we noticed that the low number of blacks in $\mat{x}_l$ (and analogously in $\mat{x}_r$) leads to some instability in the projection. This issue may occur in very particular situations where the cut happens almost in the blank area following a symbol and either there are no symbols in the sequence or the blank area is large enough so that $\mat{x}_l$ is practically blank.

\subsection{Case 4}
\begin{figure}[htb]
	\centering
	\includegraphics[width=0.6\textwidth]{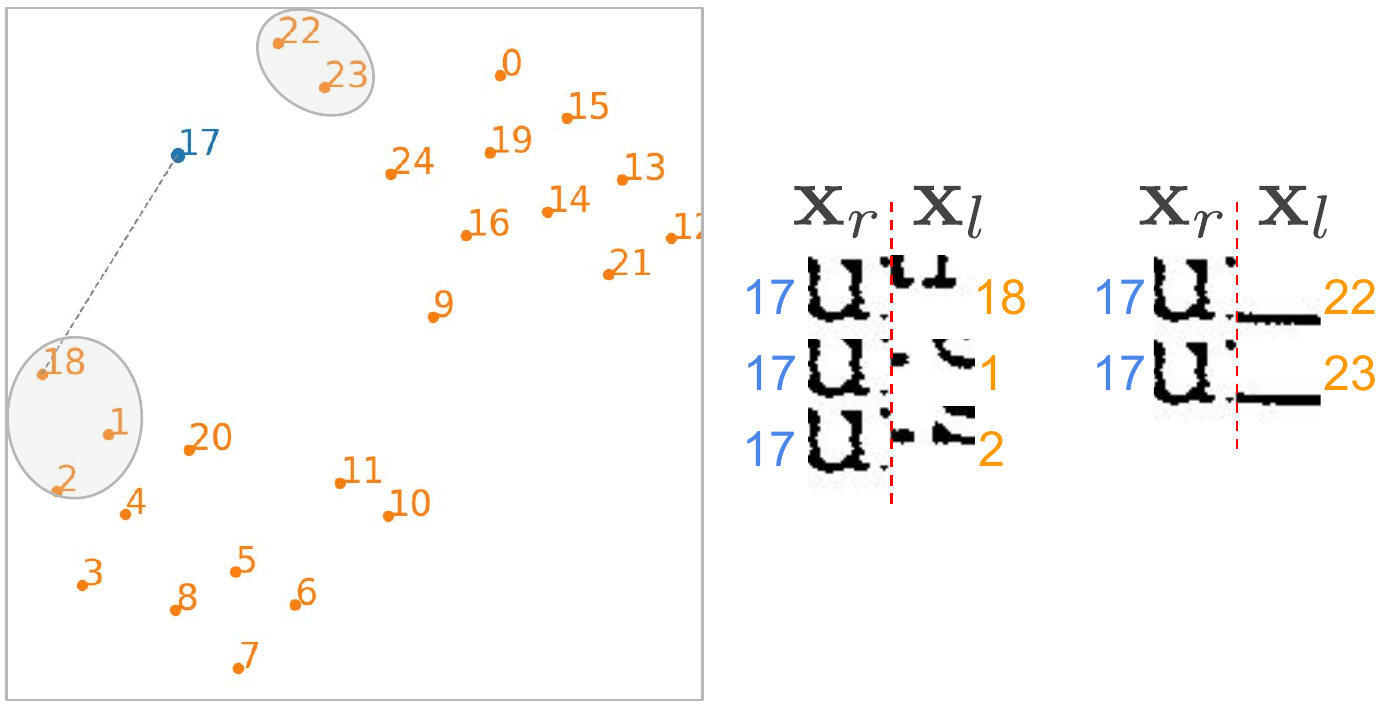}
	\caption{Case 4.} 
	\label{fig:case4}
\end{figure}

The last selected case (Figure \ref{fig:case4}) emphasizes the relevance of the vertical alignment stage of our method (Section 3.2 of the manuscript). By observing the correct pairing $(17, 18)$, it is noticeable the vertical misalignment between the shreds. The samples $22$ and $23$ are very similar, and therefore they are mapped closely in the embedding space. Also, these samples are good competitors because of the alignment with the anchor's baseline. Finally, it can be observed (as in Case 2) the clustering induced by the displaced content of $\mat{x}_l$.

\section{Sensitivity analysis w.r.t. sample size}

\begin{figure}[h]
	\centering
	\includegraphics[width=0.55\textwidth]{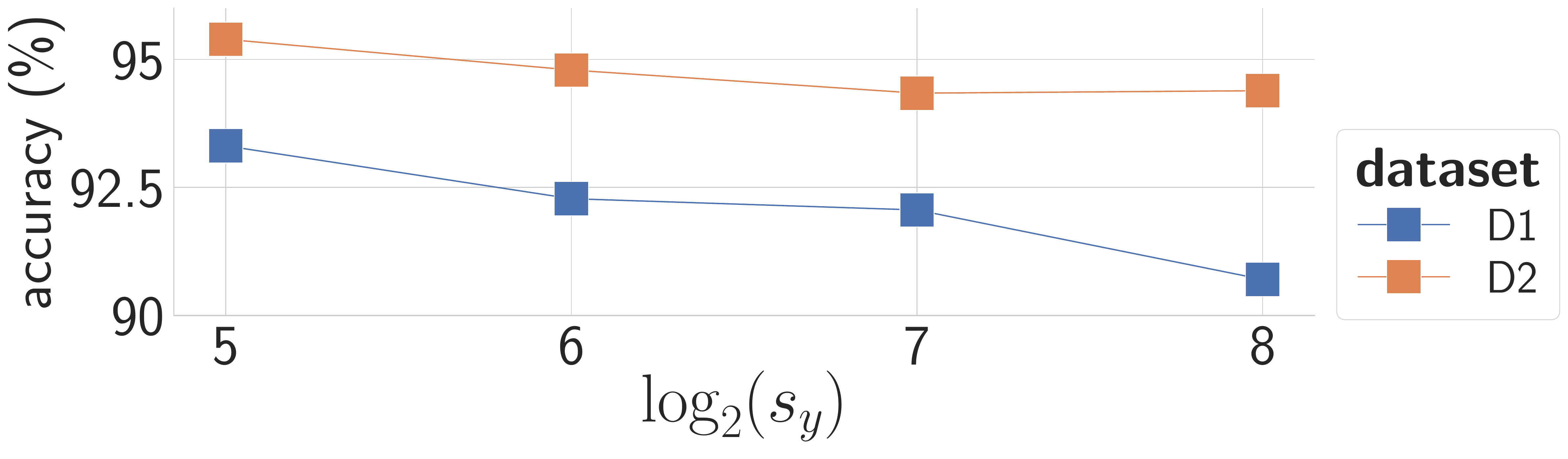}
	\caption{Reconstruction accuracy w.r.t. to the sample height ($s_y$).}
	\label{fig:ablation_input_size}
\end{figure}

As \citesupplement{paixao2018deep_app}, we use small samples ($32 \times 32$) to explore features at text line (local) level based under the assumption of weak feature correlation across text lines. In a previous investigation, we observed that the accuracy of \citesupplement{paixao2018deep_app} decreases for larger samples. This is also verified in our method when the sample height ($s_y$) is increased, as seen in Figure~\ref{fig:ablation_input_size}.

\section{Statistical test}

Considering a threshold of 5\%, the proposed method was statistically equivalent to \citesupplement{paixao2018deep_app} in D2 and superior to \citesupplement{paixao2018tifs_app} in both D1 and D2. Table~\ref{tab:t-test} shows the $p$-values of the page-wise paired $t$-test.

\begin{table}[htb]
	\centering
	\begin{adjustbox}{max width=\textwidth}
		\begin{tabular}{lccc} \toprule
			& D1 $\cup$ D2 & D1 & D2 \\
			\midrule
			\textit{Proposed} vs. \citesupplement{paixao2018deep_app} & 0.016 & 0.007 & \textbf{0.522} \\
			\textit{Proposed} vs. \citesupplement{paixao2018tifs_app} & 0.000 & 0.000 & 0.004\\
			\bottomrule
		\end{tabular}
	\end{adjustbox}
	\caption{Page-wise paired $t$-test.}
	\label{tab:t-test}
\end{table}

\nocitesupplement{*}
\bibliographystylesupplement{plain}
\bibliographysupplement{supplement}

\end{appendices}
\end{document}

%% file: secs/1_introduction.tex
% http://www.forensicsciencesimplified.org/docs/how.html
% https://books.google.com.br/books?hl=pt-BR&lr=&id=P0Z1AgAAQBAJ&oi=fnd&pg=PP1&dq=%22forensic+document+examination%22&ots=LIhHhvuWCV&sig=6nYU4Ui8chODZPE7DpQPCsdTB_M#v=onepage&q=%22forensic%20document%20examination%22&f=false
\section{Introduction}
\label{sec:intro}
% forensics => forensic science
\noindent Paper documents are of great value in forensics because they may contain supporting evidence for criminal investigation (e.g., fingerprints, bloodstains, textual information). Damage on these documents, however, may hamper or even prevent their analysis, particularly in cases of chemical destruction. Nevertheless, recent news \cite{news2018thehill} shows that documents are still being physically damaged by hand-tearing or using specialized paper shredder machines (mechanically shredding). In this context, a forensic document examiner (FDE) is typically required to reconstruct the original document for further analysis.

To accomplish this task, FDEs usually handle paper fragments (shreds) manually, verifying the compatibility of pieces and grouping them incrementally. Despite its relevance, this manual process is time-consuming, laborious, and potentially damaging to the shreds. For these reasons, research on automatic digital reconstruction has emerged since the last decade \cite{ukovich2004,justino2006}. Traditionally, hand-tearing and mechanical-shredding scenarios are addressed differently since shreds' shape tends to be less relevant in the latter. Instead, shreds' compatibility is almost exclusively determined by appearance features, such as color similarity around shreds extremities \cite{skeoch2006,marques2013}.

As with the mechanical shredding, ad hoc strategies have been also developed for binary text documents to cope with the absence of discriminative color information \cite{lin2012,sleit2013,gong2016,chen2017a}. More recently, Paix\~ao \etal \cite{paixao2018tifs} substantially improved the state-of-the-art in terms of accuracy on the reconstruction of strip-shredded text documents, i.e., documents cut in the longitudinal direction only. Nevertheless, time efficiency is a bottleneck because shreds' compatibility demands a costly similarity assessment of character shapes. In a follow-up work \cite{paixao2018deep}, the group proposed a deep learning-based compatibility measure, which improved the accuracy even further as well as the time efficiency of the reconstruction. In \cite{paixao2018deep}, shreds' compatibility is estimated pairwise by a CNN trained in a self-supervised way, learning from intact (non-shredded) documents. Human annotation is not required at any stage of the learning process. A sensitive issue, however, is that model inference is required whenever a pair of shreds has to be evaluated. Although this is not critical for a low number of shreds, scalability is compromised for a more realistic scenario comprising hundreds/thousands of shreds from different sources.

To deal with this issue, we propose a model in which the number of inferences scales linearly with the number of shreds, rather than quadratically. For that, the raw content of each shred is projected onto a space in which the distance metric is proportional to the compatibility. The projection is performed by a deep model trained using a metric learning approach. The goal of metric learning is to learn a distance function for a particular task. It has been used in several domains, ranging from the seminal work of the Siamese networks \cite{bromley1994neurips} in signature verification, to an application of the triplet loss \cite{triplet2009jmlr} in face verification \cite{facenet2015cvpr}, to the lifted structured loss \cite{lifted2016cvpr}, to the recent connection with mutual information maximization \cite{tschannen2019mutual} and many others. Unlike most of these works, however, the proposed method does not employ the same model to semantically different samples. In our case, right and left shreds are (asymmetrically) projected by two different models onto a common space. After that, the distances between the right and left shreds are measured, the compatibility matrix is built and passed on to the actual reconstruction. To enable fair comparisons, the actual reconstruction was performed by coupling methods for compatibility evaluation to an external optimizer. The experimental results show that our method achieves accuracy comparable to the state-of-the-art ($97.22\%$) while taking only 3.73 minutes to reconstruct 20 mixed pages with 505 shreds compared to 1 hour and 20 minutes of \cite{paixao2018deep}, i.e., a speed-up of $\approx22$ times.

In summary, the main contributions of our work are:
\begin{enumerate}
    \item This work proposes a compatibility evaluation method leveraging metric learning and the asymmetric nature of the problem;
    \item The proposed method does not require manual labels (trained in a self-supervised way) neither real data (the model is trained with artificial data);
    \item The experimental protocol is extended from a single-page to a more realistic and time demanding scenario with a multi-page multi-document reconstruction;
    \item Our proposal scales the inference linearly rather than quadratically as in the current state-of-the-art, achieving a speed-up of $\approx22$ times for 505 shreds, and even more for more shreds.
\end{enumerate}

%% file: secs/2_problem.tex
\section{Problem Definition}
\label{sec:problem}

\begin{figure}[t]
    \centering
    \includegraphics[width=\columnwidth]{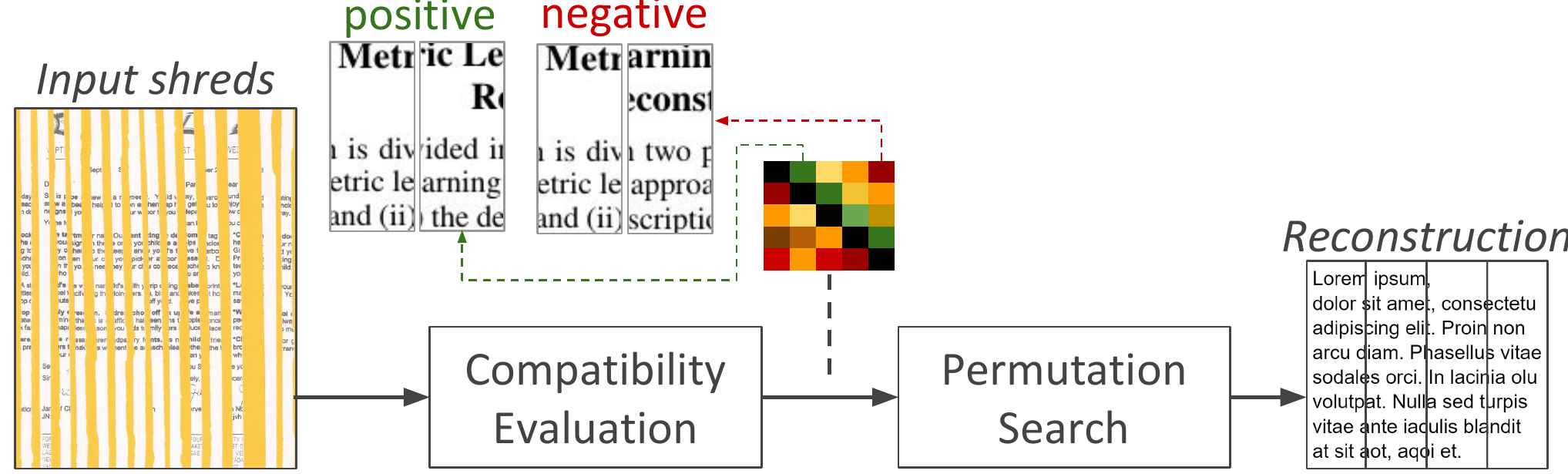} 
    \caption{Classical approach for automatic document reconstruction. Shreds' compatibility is evaluated pairwise and then an optimization search process is conducted (based on the compatibility values) in order to find the shreds' permutation that best represents the original document.}
    \label{fig:problem}
\end{figure}

For simplicity of explanation, let us first consider the scenario where all shreds belong to the same page: single-page reconstruction of strip-shredded documents.
Let $\mathcal{S} = \{s_1, s_2, \ldots, s_{n}\}$ denote the set of $n$ shreds
resulting from longitudinally shredding (strip-cut) a single page. Assume that the indices determine the ground-truth order of the shreds: $s_1$ is the leftmost shred, $s_2$ is the right neighbor of $s_1$, and so on. A pair $(s_i, s_j)$ -- meaning $s_j$ placed right after $s_i$ -- is said to be ``positive'' if $j = i + 1$, otherwise it is ``negative''. A solution of the reconstruction problem can
be represented as a permutation $\pi_{\mathcal{S}} = (s_{\pi_1}, s_{\pi_2}, \ldots, s_{\pi_n})$ of $\mathcal{S}$.
A perfect reconstruction is that for which $\pi_i = i$, for all $i = 1, 2, \ldots, n$.

Automatic reconstruction is classically formulated as an optimization problem
\cite{prandtstetter2008,morandell2008} whose objective function derives from pairwise compatibility (Figure~\ref{fig:problem}).
Compatibility or cost, depending on the perspective, is given by a
function $c: \mathcal{S} \times \mathcal{S} \to \mathbb{R}$ that quantifies the (un)fitting of two shreds when placed side-by-side (order matters). Assuming a cost interpretation, $c(s_i, s_j)$, $i \neq j$, denotes the
cost of placing $s_j$ to the right of $s_i$. In theory, $c(s_i, s_j)$ should be low when $j = i + 1$ (positive pair),
and high for other cases (negative pairs). Typically, $c(s_i, s_j) \neq c(s_j, s_i)$ due to the asymmetric nature of the
reconstruction problem.

The cost values are the inputs for a search procedure that aims to find the optimal permutation $\pi_\mathcal{S}^{*}$, i.e., the arrangement of the shreds that best resembles the original document. The objective function $C$ to be minimized is the accumulated pairwise cost computed only for
consecutive shreds in the solution:
\begin{equation}
\label{eq:objective}
C(\pi_{\mathcal{S}}) = \sum_{i=1}^{n - 1}{c(s_{\pi_i}, s_{\pi_{i + 1}})}.
\end{equation}
The same optimization model can be applied in the reconstruction of several shredded pages
from one or more documents (multi-page reconstruction).
In a stricter formulation, a perfect solution in this scenario can be represented by a sequence of shreds which respects the ground-truth order in each page, as well as the expected
order (if any) of the pages themselves. If page order is not relevant (or does not apply),
the definition of a positive pair of shreds can be relaxed, such that a pair $(s_i, s_j)$ is also positive if $s_i$ and $s_j$ are, respectively, the last and first shreds of different pages, even for $j \neq i + 1$.
The optimization problem of minimizing Equation~\ref{eq:objective} has been extensively investigated in literature, mainly using
genetic algorithms \cite{biesinger2013,xu2014,ge2015reconstructing,gong2016}
and other metaheuristics \cite{prandtstetter2009meta,schauer2010,badawy2018discrete}. The focus of this
work is, nevertheless, on the compatibility evaluation between shreds (i.e., the function $c$), which is critical to lead the search towards accurate reconstructions.

\begin{figure}[t]
    \centering
    \includegraphics[width=\columnwidth]{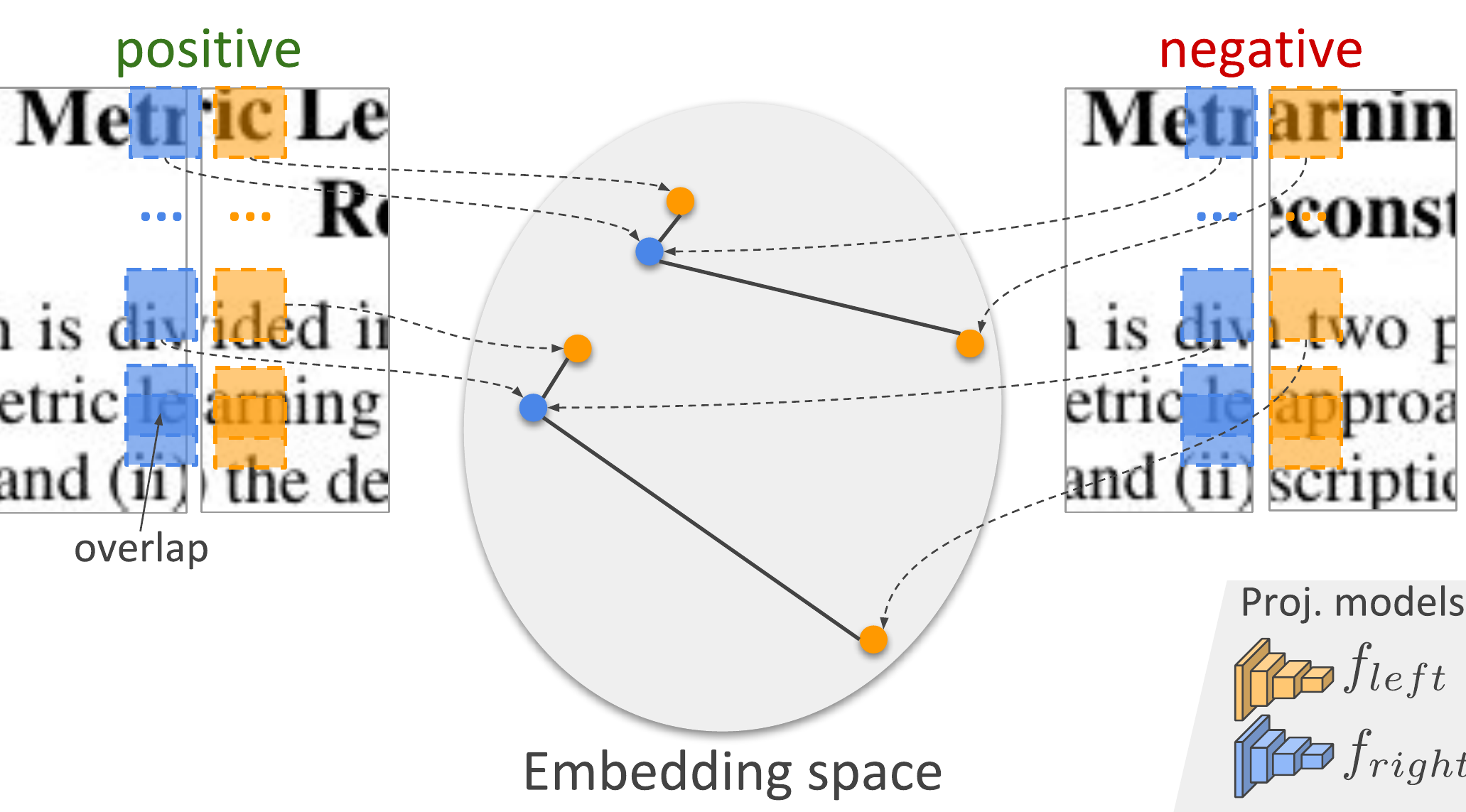} 
    \caption{Metric learning approach for shreds' compatibility evaluation. Embeddings generated
        from compatible regions are expected to be closer in the embedding space, whereas those from non-fitting regions are expected to be mapped far from each other.}
    \label{fig:embedding}
\end{figure}

To address text documents, literature started to evolve from the application of pixel-level
similarity metrics \cite{balme2007,gong2016,marques2013}, which are fast but inaccurate, towards
stroke continuity analysis \cite{phienthrakul2015,guo2015} and symbol-level matching
\cite{xing2017a,paixao2018tifs}. Strokes continuity across shreds, however, cannot be ensured since physical shredding
damages the shreds' borders. Techniques based on symbol-level features, in turn, tend to be more robust. However they may struggle to segment symbols in complex documents, and to cope efficiently with the wide variability of symbols' shape and size. These issues have been addressed in \cite{paixao2018deep}, wherein deep learning has been successfully used for accurate reconstruction of strip-shredded documents. Nonetheless, the large number of network inferences required for compatibility assessment hinders scalability for multi-page reconstruction.

This work addresses precisely the scalability issue. Although our self-supervised approach shares some similarities with their work, the training paradigm is completely distinct since the deep models here do not provide compatibility (or cost) values. Instead, deep models are used to convert pixels into embedding representations, so that a simple distance metric can be applied to measure shreds' compatibility. This is better detailed in the next section.

%% file: secs/3_proposed.tex
% https://math.stackexchange.com/questions/2788631/mathematical-notation-for-matrix-slicing
% Reference book for notation - https://www.deeplearningbook.org/contents/linear_algebra.html
\section{Compatibility Evaluation via Deep Metric Learning}
\label{sec:proposed}

The general intuition behind the proposed approach for compatibility evaluation is illustrated in Figure~\ref{fig:embedding}. The underlying assumption is that two side-by-side shreds are globally compatible if they locally fit each other along the touching boundaries. The local approach relies on small samples (denoted by $\mathbf{x}$) cropped from the boundary regions. Instead of comparing pixels directly, the samples are first converted to an intermediary representation (denoted by $\mathbf{e}$) by projecting them onto a common embedding space $\mathbb{R}^d$. Projection is accomplished by two models (CNNs): $f_{left}$ and $f_{right}$, $f_{\bullet} : \mathbf{x} \mapsto \mathbf{e}$, specialized on the left and right boundaries, respectively. 

Assuming that these models are properly trained, boundary samples (indicated by the orange and blue regions in Figure~\ref{fig:embedding}) are then projected, so that embeddings generated from compatible regions (mostly found on positive pairings) are expected to be closer in this metric space, whereas those from non-fitting regions should be farther apart. Therefore, the global compatibility of a pair of shreds is measured in function of the distances between corresponding embeddings. More formally, the cost function in Equation~\ref{eq:objective} is such that:
\begin{equation}
\label{eq:cost}
c(s_i, s_j) \propto \phi(\mathbf{e}_i, \mathbf{e}_j),
\end{equation}
where $\mathbf{e}_{\bullet}$ represents the embeddings associated with the shred $s_{\bullet}$, and $\phi$ is a distance metric (e.g., Euclidean).

The interesting property of this evaluation process is that the projection step can be decoupled from the distance computation. In other words, the process scales linearly since each shred is processed once by each
model, and pairwise evaluation can be performed with the embeddings produced. Before diving into the details of the evaluation, we first describe the self-supervised learning of these models. Then, a more in-depth view of evaluation will be presented, including the formal definition of a cost function that composes the objective function in Equation~\ref{eq:objective}.

\subsection{Learning Projection Models}

For producing the shreds' embeddings, the models $f_{left}$ and $f_{right}$ are trained simultaneously with small $s \times s$ samples. The two models have the same fully convolutional architecture: a base network for feature extraction appended with a convolutional layer. The added layer is intended to work as a fully connected layer when the base network is fed with $s \times s$ samples. Nonetheless, weight sharing is disabled since models specialize on different sides of the shreds, hence deep asymmetric metric learning. The base network comprises the first three convolutional blocks of SqueezeNet \cite{iandola2016squeezenet} architecture (i.e., until the \emph{fire3} block).

SqueezeNet has been effectively used in distinguishing between valid and invalid symbol patterns in the context of compatibility evaluation \cite{paixao2018deep}. Nevertheless, preliminary evaluations have shown that the metric learning approach is more effective with shallower models, which explains the use of only the first three blocks. For projection onto $\mathbb{R}^d$ space, a convolutional layer with $d$ filters of dimensions
$s/4 \times s/4$ (base network's dimensions when fed with $s\times s$ samples)
and sigmoid activation was added to the base network.

\begin{figure}[t]
	\centering
	\includegraphics[width=\columnwidth]{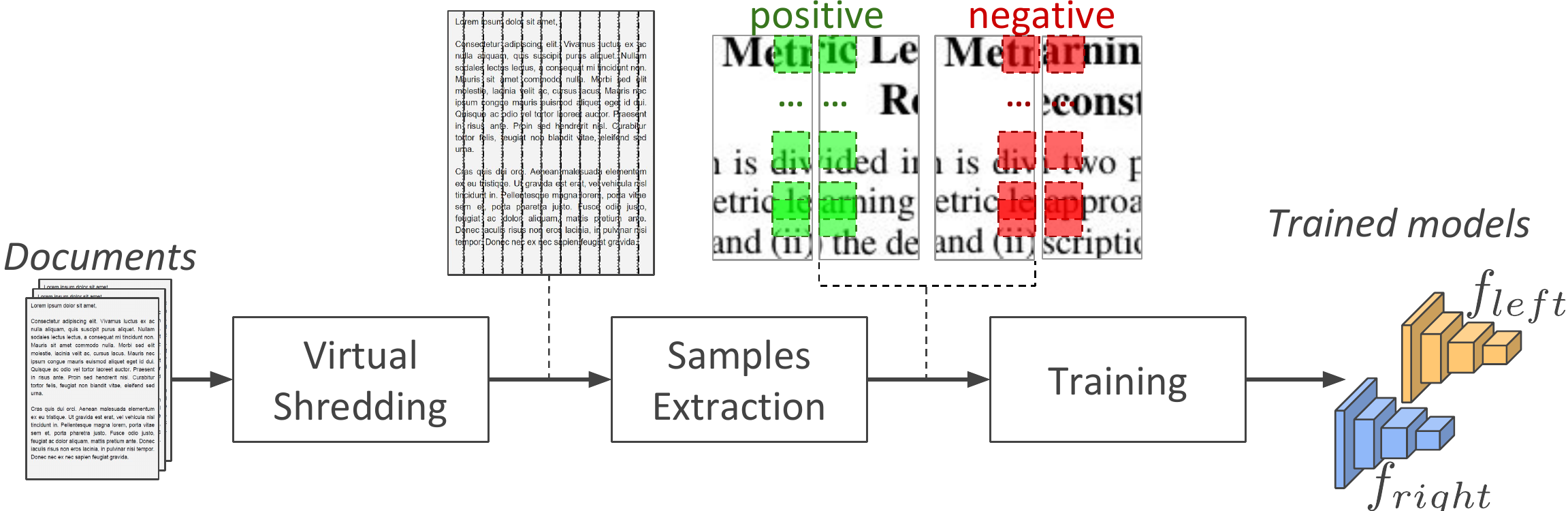}
	\caption{Self-supervised learning of the models with samples extracted from digital documents.}
	\label{fig:trainingOverview}
\end{figure}

Figure~\ref{fig:trainingOverview} outlines the self-supervised learning of the models with samples extracted from digital documents. First, the shredding process is simulated so that the digital documents are cut into equally shaped rectangular ``virtual'' shreds.
Next, shreds of the same page are paired side-by-side and sample pairs are extracted top-down along the touching edge: one sample from the $s$ rightmost pixels of the left shred (r-sample), and the other from the $s$ leftmost pixels of the right shred (l-sample). Since shreds' adjacency relationship is provided for free with virtual shredding, sample pairs can be automatically labeled as ``positive'' (green boxes) or ``negative'' (red boxes). Self-supervision comes exactly from the fact that labels are automatically acquired by exploiting intrinsic properties of the data.

\begin{figure}[t]
	\centering
	\includegraphics[width=\columnwidth]{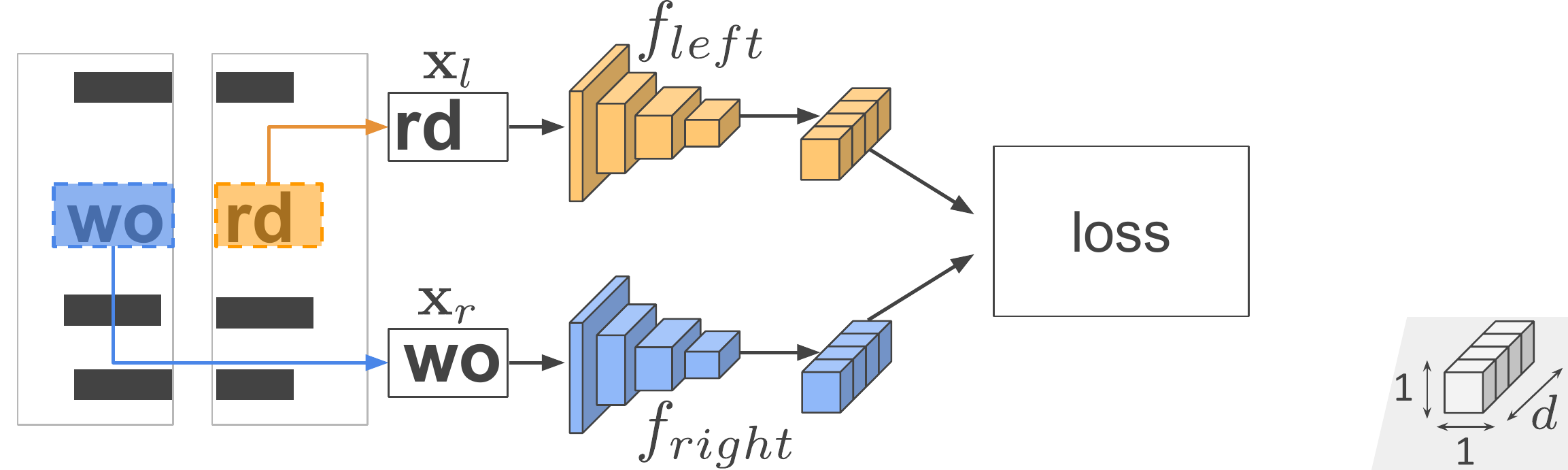} 
	\caption{Learning projection models for shreds' compatibility evaluation.
		The models are jointly trained with sample pairs guided by the
		contrastive loss function. The input vectors for the loss function are encoded as $1 \times 1 \times d$ tensors.}
	\label{fig:loss}
\end{figure}

Training data comprise tuples $(\mat{x}_r, \mat{x}_l, y)$, where $\mat{x}_r$ and $\mat{x}_l$ denote, respectively, the r- and l-samples of a sample pair, and $y$ is the associated ground-truth label: $y=1$ if the sample pair is positive, and $y=0$, otherwise. Training is driven by the contrastive loss function \cite{chopra2005learning}:
\begin{equation}
	\label{eq:loss}
	\begin{split}
		\mathcal{L}(f_{left}, f_{right}, \mat{x}_l, \mat{x}_r, y) & = \\
		\frac{1}{2}\{(1 - y) \cdot dist^2 + & y \cdot [\max(0, m - dist)]^2\},
	\end{split}
\end{equation}
where $dist = {\left \|f_{left}(\mathbf{x}_l) - f_{right}(\mathbf{x}_r)\right \|}_2$, and $m$ is the margin parameter. For better understanding, an illustration is provided in Figure~\ref{fig:loss}. The models handle a positive sample pair that, together, composes the pattern ``word''. Since it is positive ($y=1$), the loss value would be low if the resulting embeddings are close in $\mathbb{R}^d$, otherwise, it would be high. Note that weight-sharing would result in the same loss value for the swapped samples (pattern ``rdwo''), which is undesirable for the reconstruction application. Implementation details of the sample extraction and training procedure are described in experimental methodology (\sect{sec:implementation}).

%%%%%%%%%%%%%%%%%%%%%%%%%%%%%%%%%%%%%%%%%%
\subsection{Compatibility Evaluation}
\label{sec:evaluation}
\begin{figure}[t]
	\centering
	\includegraphics[width=\columnwidth]{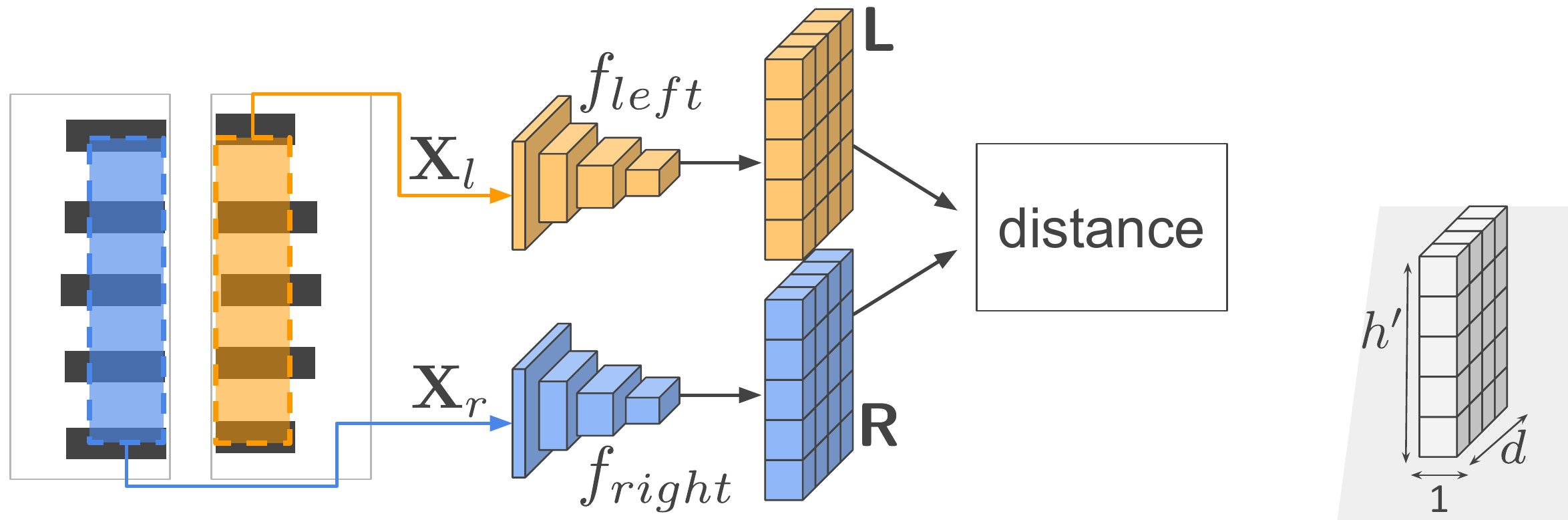} 
	\caption{Compatibility evaluation of a pair of shreds. Local embeddings, represented
		by the $h' \times 1 \times d$ tensors $\ten{L}$ and $\ten{R}$, are extracted along the boundary regions.
		Compatibility is a real value given by the squared Euclidean distance between $\ten{L}$ and $\ten{R}$ (computed over the flattened tensors).}
	\label{fig:distance}
\end{figure}

%%%%%%%%%%%%%%%%%%%%%%%%%%%%%%%%%%%%%%%%%%
In compatibility evaluation, shreds' embedding and distance computation are two decoupled steps. Figure~\ref{fig:distance} presents a joint view of these two steps for better understanding of the model's operation. Strided sliding window is implicitly performed by the fully convolutional models. To accomplish this, two vertically centered $h \times s$ regions of interest are
cropped from the shreds' boundaries ($s$ is the sample size): $\mat{X}_r$, comprising the
$s$ rightmost pixels of the left shred, and $\mat{X}_l$, comprising the $s$ leftmost pixels
of the right shred. Inference on the models produces $h' \times 1 \times d$ feature volumes represented by the tensors $\ten{L}=f_{left}(\mat{X}_l)$ (l-embeddings) and $\ten{R}=f_{right}(\mat{X}_r)$ (r-embeddings). The $h'$ rows from the top to the bottom of the tensors represent exactly the top-down sequence of $d$-dimensional local embeddings illustrated in Figure~\ref{fig:embedding}.

If it is assumed that vertical misalignment among shreds is not significant, compatibility could be obtained by simply computing ${\left \|\ten{R} - \ten{L} \right \|}_2$. For a more robust definition, shreds can be vertically ``shifted'' in the image domain to account for misalignment \cite{paixao2018deep}. Alternatively, we propose to shift the tensor $\ten{L}$ ``up'' and ``down'' $\delta$ units (limited to $\delta_{max}$) in order to determine the best-fitting pairing, i.e., that which yields the lowest cost. This formulation helps to save time since it does not require new inferences on the models. Given a tensor $\ten{T} = {(T_{i,j,k})}_{h' \times 1 \times d}$,
let $\ten{T}_{a:b} = {(T_{i,j,k})}_{a \leq i \leq b,~j=1,~1 \leq k \leq d}$ denote
a vertical slice from row $a$ to $b$. Let $\ten{R}^{(i)}$ and $\ten{L}^{(j)}$ represent, respectively, the r- and l-embeddings for a pair of shreds $(s_i, s_j)$. When shifts are restricted to the upward direction, compatibility is defined by the function:
\begin{equation}
\label{eq:up}
c_{\uparrow}(s_i, s_j) = \min_{0 \leq \delta \leq \delta_{max}}{\left \|\ten{R}^{(i)}_{1 : 1 + n_{rows}} - \ten{L}^{(j)}_{1 + \delta : 1 + n_{rows} + \delta}\right \|}_2,
\end{equation}
where $n_{rows} = h' - \delta_{max}$ is the number of rows effectively used for distance computation. Analogously, for the downward direction:
\begin{equation}
\label{eq:down}
c_{\downarrow}(s_i, s_j) = \min_{0 \leq \delta \leq \delta_{max}}{\left \|\ten{R}^{(i)}_{1 + \delta : 1 + n_{rows} + \delta} - \ten{L}^{(j)}_{1 : 1 + n_{rows}}\right \|}_2
\end{equation}
Finally, the proposed cost function is a straightforward combination of Equations \eqref{eq:up} and \eqref{eq:down}:
\begin{equation}
\label{eq:compatibility}
c(s_i, s_j) = \min(c_{\uparrow}(s_i, s_j), c_{\downarrow}(s_i, s_j)).
\end{equation}
Note that, if $\delta_{max}$ is set to $0$ (i.e., shifts are not allowed), then $n_{rows} = h'$,
therefore:
\begin{equation}
\label{eq:compatibility_null}
c(i, j) = c_{\uparrow}(i, j) = c_{\downarrow}(i, j) = {\left \|\ten{R}^{(i)} - \ten{L}^{(j)} \right \|}_2.
\end{equation}

%% file: secs/4_experimental.tex
\section{Experimental Methodology}
\label{sec:experimental}
 
The experiments aim to evaluate the accuracy and time performance of the proposed method, as well as to compare with the literature in document reconstruction focusing on the deep learning method proposed by Paix\~ao \etal \cite{paixao2018deep} 
(hereafter referred to as \base). For this purpose, we followed the basic protocol proposed in \cite{paixao2018tifs} in which the methods are coupled to an exact optimizer and tested on two datasets (D1 and D2).
Two different scenarios are considered here: single- and multi-page reconstruction.

\subsection{Evaluation Datasets}

\paragraph{D1.} Produced by Marques and Freitas \cite{marques2013}, it comprises $60$ shredded pages scanned at $300$ dpi. Most pages are from academic documents (e.g., books and thesis), part of such pages belonging to the same document. Also, $39$ instances have only textual content, whereas the other $21$ have some graphic element (e.g., tables, diagrams, photos). Although a real machine (Cadence FRG712) has been used, the shreds present almost uniform dimensions and shapes.
Additionally, the text direction is nearly horizontal in most cases.

\paragraph{D2.} This dataset was produced by Paix\~ao \etal \cite{paixao2018tifs} and comprises $20$ single-page documents (legal documents and business letters) from the ISRI-Tk OCR collection \cite{nartker2005}. The pages were shredded with a Leadership 7348 strip-cut machine and their respective shreds were arranged side-by-side onto a support yellow paper sheet, so that they could be scanned at once and, further, easily segmented from background. In comparison to D1, the shreds of D2 have less uniform shapes and their borders are significantly more damaged due to the machine blades wear. Besides, the handling of the shreds before scanning caused slight rotation and (vertical) misalignment between the shreds. These factors render D2 as a more realistic dataset compared to D1.

\subsection{Accuracy Measure}

Similar to the neighbor comparison measure \cite{andalo2017}, the accuracy of a solution is defined here as the fraction of adjacent pairs of shreds which are ``positive''. For multi-reconstruction, the relaxed definition of ``positive'' is assumed (as discussed in \sect{sec:problem}), i.e., the order in which the pages appear is irrelevant. More formally, let $\pi_{\mathcal{S}} = (s_{\pi_1}, s_{\pi_2}, \ldots, s_{\pi_n})$ be a solution for the reconstruction
problem for a set of shreds $\mathcal{S}$. Then, the accuracy of $\pi_{\mathcal{S}}$ is calculated as
\begin{equation}
\label{eq:accuracy}
\operatorname{accuracy}(\pi_{\mathcal{S}}) = \frac{1}{n - 1}\sum_{i=1}^{n-1}{\bf 1}[(s_{\pi_i}, s_{\pi_{i+1}}) \text{ is positive}],
\end{equation}
where $\operatorname{\bf 1}[\cdot]$ denotes the 0-1 indicator function.

\subsection{Implementation Details}
\label{sec:implementation}

\paragraph{Sample Extraction.} Training data consist of $32 \times 32$ samples extracted from $100$ binary documents (forms, emails, memos, etc.) scanned at $300$ dpi of the IIT-CDIP Test Collection 1.0 \cite{lewis2006building}. For sampling, the pages are split longitudinally into $30$ virtual shreds (amount estimated for the usual A4 paper shredders). Next, the shreds are individually thresholded with Sauvola's algorithm \cite{sauvola2000adaptive} to cope with small fluctuations in pixel values of the original images. Sample pairs are extracted page-wise, which means that the samples in a pair come from the same document. The extraction process starts with adjacent shreds in order to collect positive sample pairs (limited to $1{,}000$ pairs per document). Negative pairs are collected subsequently, but limited to the number of positive pairs. During extraction, the shreds are scanned from top to bottom, cropping samples every two pixels. Pairs with more than $80\%$ blank pixels are considered ambiguous, and then they are discarded for future training. Finally, the damage caused by mechanical shredding is roughly simulated with the application of salt-and-pepper random noise on the two rightmost pixels of the r-samples, and on the two leftmost pixels of the l-samples.

\paragraph{Training.} The training stage leverages the sample pairs extracted from the collection of $100$ digital documents. From the entire collection, the sample pairs of $10$ randomly picked documents are reserved for validation where the best-epoch model should be selected. By default, the embeddings dimension $d$ is set to $128$. The models are trained from scratch (i.e., the weights are randomly initialized) for $100$ epochs using the stochastic gradient descent (SGD) with a learning rate of $10^{-1}$ and mini-batches of size $256$. After each epoch, the models' state is stored, and the training data are shuffled for the new epoch (if any). The best-epoch model selection is based on the ability to project positive pairs closer in the embedding space, and negative pairs far. This is quantified via the standardized mean difference (SMD) measure \cite{cohn1988statistical} as follows: for a given epoch, the respective $f_{left}$ and $f_{right}$ models are fed with the validation sample pairs and the distances among the corresponding embeddings are measured. Then, the distance values are separated into two sets: ${dist}^+$, comprising distances calculated for positive pairs, and ${dist}^-$, for negative ones.
Ideally, the difference between the mean values of the two sets should be high, while the standard deviations within the sets should be low. Since these assumptions are addressed in SMD, the best $f_{left}$ and $f_{right}$ are taken as those which maximize $\operatorname{SMD}({dist}^+, {dist}^-)$.

\subsection{Experiments}
\label{sec:experiments}

The experiments rely on the trained models $f_{left}$ and $f_{right}$, as well as on the \baseposs deep model. The latter was retrained (following the procedure described in \cite{paixao2018deep}) on the CDIP documents to avoid training and testing with documents of the same collection (ISRI OCR-Tk). In practice, no significant change was observed in the reconstruction accuracy with this procedure.

The shreds of the evaluation datasets were also binarized \cite{sauvola2000adaptive} to keep consistency with training samples. The default parameters of the proposed method includes $d=128$ and $\delta_{max}=3$. Non-default assignments are considered in two of the three conducted experiments, as better described in the following subsections.

\paragraph{Single-page Reconstruction.} This experiment aims to show whether the proposed method is able to individually reconstruct pages with accuracy similar to \base, and how the time performance of both methods is affected when the vertical shift functionality is enabled since it increases the number of pairwise evaluations. To this intent, the shredded pages of D1 and D2 were individually reconstructed with the proposed
and \base methods, first using their default configuration, and after disabling the vertical shifts (in our case, it is equivalent to set $\delta_{max}=0$). Time and accuracy were measured
for each run. For a more detailed analysis, time was measured for each reconstruction stage: projection (pro) -- applicable only for the proposed method --, pairwise compatibility evaluation (pw), and optimization search (opt).

\paragraph{Multi-page Reconstruction.} This experiment focuses on the scalability with respect to time while increasing the number of shreds in multi-page reconstruction. In addition to the time performance, it is essential to confirm whether the accuracy of both methods remains comparable. Rather than individual pages, there are two large reconstruction instances in this experiment: the $1{,}370$ mixed shreds of D1 and the $505$ mixed shreds of D2. Each instance was reconstructed with the proposed and \base methods, but now only with their default configuration (i.e., vertical shifts enabled). Accuracy and time (segmented by stage) were measured. Additionally, time processing was estimated for different instance sizes based on the average elapsed time observed for D2.

\paragraph{Sensitivity Analysis.} The last experiment assesses how the proposed method is affected (time and accuracy) by testing with different embedding dimensions ($d$): $2, 4, 8, \ldots, 512$. Note that this demands
the retraining of $f_{left}$ and $f_{right}$ for each $d$. After training, the D1 and D2 instances were individually reconstructed, and then accuracy and time processing were measured.

\subsection{Experimental Platform}

The experiments were carried out in an Intel Core i7-4770 CPU @ 3.40GHz with 16GB of RAM running Linux Ubuntu 16.04, and equipped with a TITAN X (Pascal) GPU with 12GB of memory. Implementation\footnote{\texttt{https://github.com/thiagopx/deeprec-cvpr20.}} was written in Python 3.5 using Tensorflow for training and inference, and OpenCV for basic image manipulation.

%% file: secs/5_results.tex
\section{Results and Discussion}
\label{sec:results}

\begin{table}[t]
    \centering
    \normalsize
    \begin{adjustbox}{max width=\columnwidth}
    \begin{tabular}{llll} \toprule
        Method & D1 $\cup$ D2 & D1 & D2\\
        \midrule
        \textbf{Proposed} & \textbf{93.71 $\pm$ 11.60} & \textbf{93.14 $\pm$ 12.93} & \textbf{95.39 $\pm$ 6.02} \\ 
        Paix\~ao-b \cite{paixao2018deep} & 96.28 $\pm$ 5.15 & 96.78 $\pm$ 4.44 & 94.78 $\pm$ 6.78 \\
        Paix\~ao \etal \cite{paixao2018tifs} & 74.85 $\pm$ 22.50 & 71.85 $\pm$ 23.14 & 83.83 $\pm$ 18.12\\ 
        Marques and Freitas \cite{marques2013} & 23.90 $\pm$ 17.95 & 29.18 $\pm$ 17.43 & 8.05 $\pm$ 6.60\\ 
        \bottomrule
    \end{tabular}
    \end{adjustbox}
    \caption{Single-page reconstruction performance: average accuracy $\pm$ standard deviation (\%).}
    \label{tab:rec}
\end{table}

\subsection{Single-page Reconstruction}

\begin{figure}[t]
    \centering
    \includegraphics[width=\columnwidth]{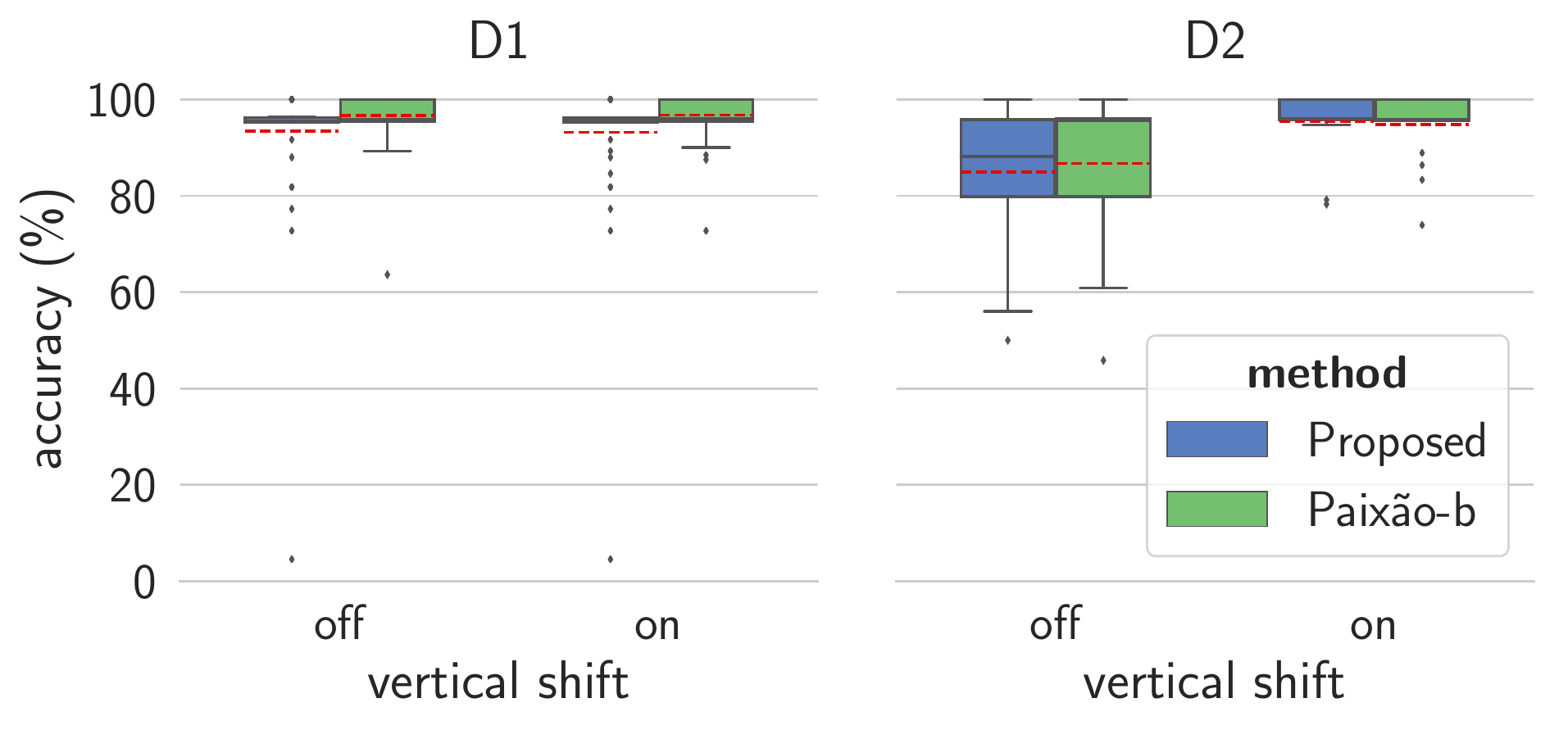} 
    \caption{Accuracy distribution for single-page reconstruction with the proposed and \base methods. Accuracies are calculated document-wise and the average values are represented by the red dashed lines.}
    \label{fig:exp1}
\end{figure}

A comparison with the literature for single-page reconstruction of strip-shredded documents is summarized in the Table~\ref{tab:rec}. Given the clear improvement in the performance, the following discussions will focus on the comparison with \cite{paixao2018deep}. The box-plots in Figure~\ref{fig:exp1} show the accuracy distribution obtained with both the proposed method and \base for single-page reconstruction.
Likewise \cite{paixao2018deep}, we also observe that vertical shifts affect only D2 since the D1's shreds are practically aligned (vertical direction). The methods did not present significant difference in accuracy for the dataset D2. For D1, however, \base slightly outperformed ours: the proposed method with default configuration (vertical shift ``on'') yielded accuracy of $93.14 \pm 12.88\%$ (arithmetic mean $\pm$ standard deviation), while \base achieved $96.78 \pm 4.44\%$. The higher variability in our approach is mainly explained by the presence of documents with large areas covered by filled graphic elements, such as photos and colorful diagrams (which were not present in the training). By disregarding these cases (12 in a total of 60 samples), the accuracy of our method increases to $95.88\%$, and the standard deviation drops to $3.84\%$.

\begin{figure}[t]
    \centering
    \includegraphics[width=\columnwidth]{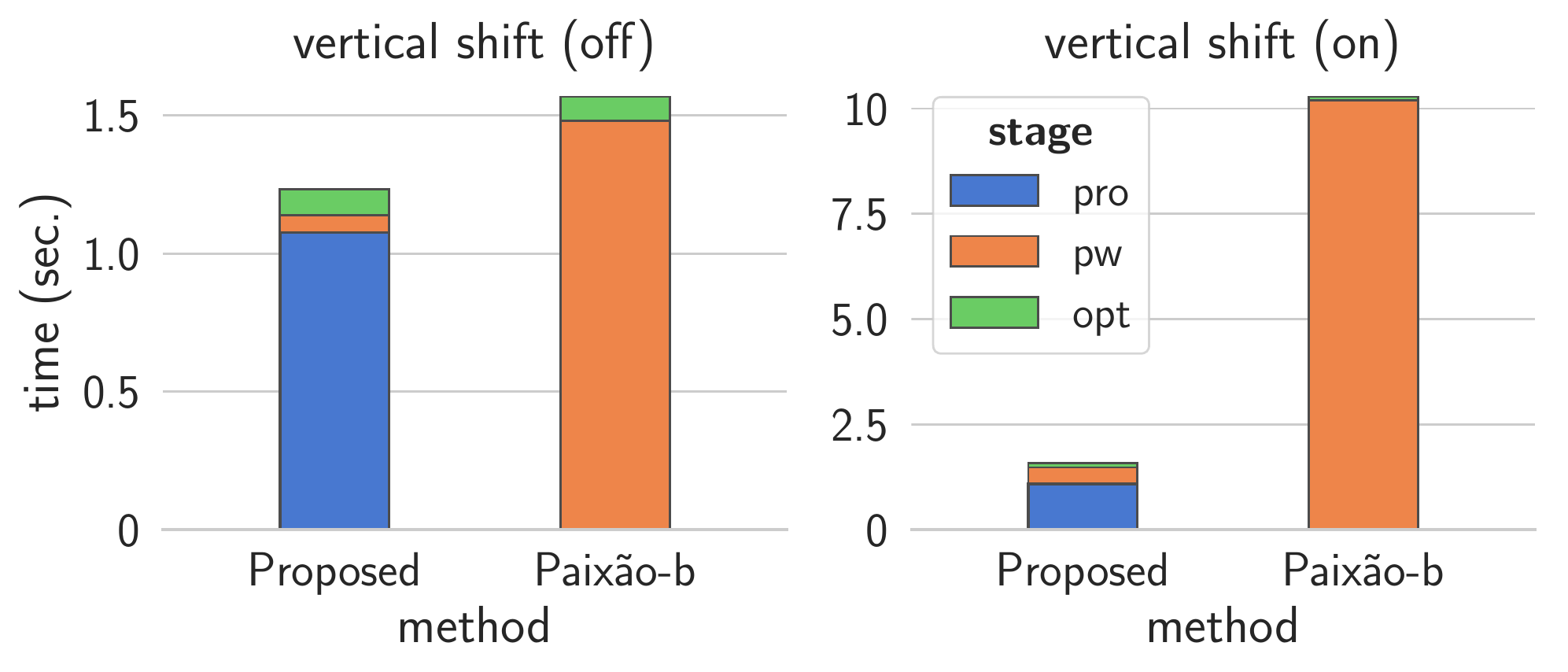} 
    \caption{Time performance for single-page reconstruction. The stacked bars represent the average elapsed time for each reconstruction stage: projection (pro), pairwise compatibility evaluation (pw), and optimization search (opt).}
    \label{fig:exp1_time}
\end{figure}

Time performance is shown in Figure~\ref{fig:exp1_time}. The stacked bars represent the average elapsed time in seconds (s) for each reconstruction stage: projection (pro), pairwise compatibility
evaluation (pw), and optimization search (opt). With vertical shift disabled (left chart), the proposed method spent much more time producing the embeddings ($1.075$s) than in pairwise evaluation ($0.063$s) and optimization ($0.092$s). Although \base does not have the cost of embedding projection, pairwise evaluation took $1.481$s, about $23$ times the time elapsed in the same stage in our method. This difference becomes more significant (in absolute values) when the number of pairwise evaluations increases, as it can be seen with the enabling of vertical shifts (right chart). In this scenario, pairwise evaluation took $0.389$s in our method, against the $10.197$s spent in \base (approx. $26$ times slower). Including the execution time of the projection stage, our approach yielded a speed-up of almost $7$ times for compatibility evaluation. Note that, without vertical shifts, the accuracy of \base would drop from $94.77\%$ to $86.74\%$ in D2.

\begin{figure*}[t]
    \centering
    \includegraphics[width=\textwidth]{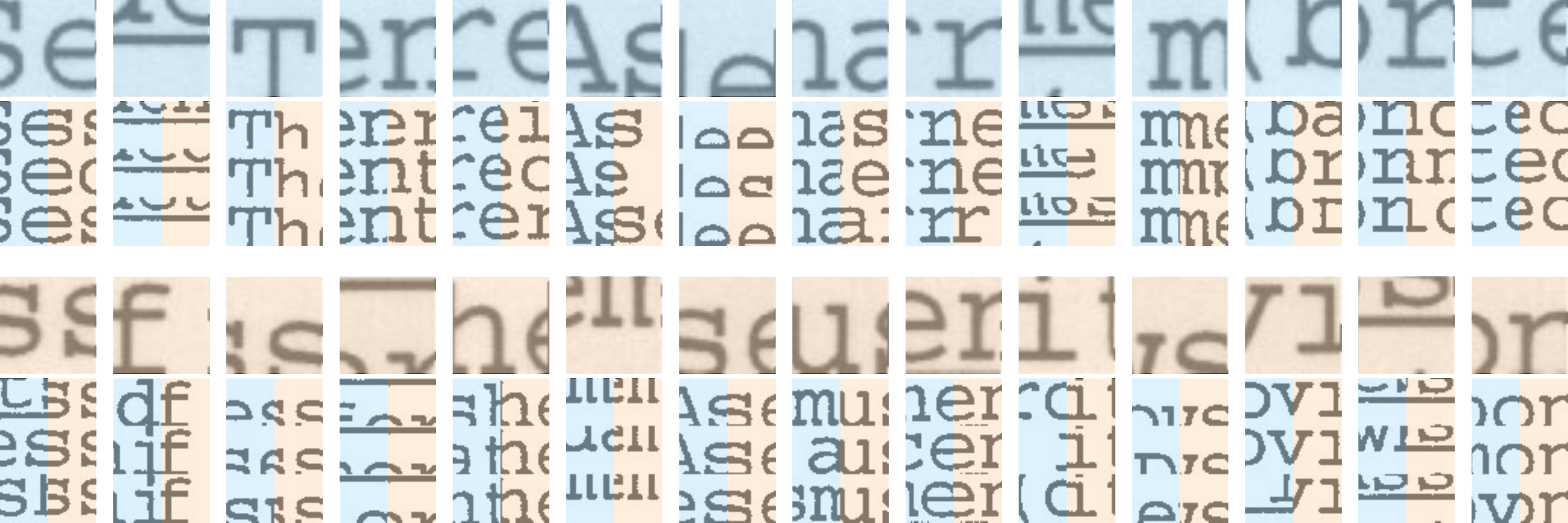} 
    \caption{Local samples nearest neighbors. In the top row, the largest square is the ``query'' sample (before binarization) followed, below, by its binary version and its three nearest neighbors side-by-side (with the closest in the top row). The blue and orange samples were projected by $f_{right}$ and $f_{left}$, respectively. The bottom row shows some examples in which the ``query'' is projected by the $f_{left}$ instead.}
    \label{fig:placeholder}
\end{figure*}

Finally, we provide an insight into what the embedding space might look like by showing a local sample and its three nearest neighbors. As shown in Figure~\ref{fig:placeholder}, the models tend to form pairs that resemble something realistic. It is worth noting that the samples are very well aligned vertically, even in cases where the sample is shifted slightly to the top or bottom and the letters are appearing only in half (see more samples in the Supplementary Material).

\subsection{Multi-page Reconstruction}

For multi-page reconstruction, the proposed method achieved $94.81\%$ and $97.22\%$ of accuracy for D1 and D2, respectively, whereas \base achieved $97.08\%$ and $95.24\%$. Overall, both methods yielded high-quality reconstructions with low difference in accuracy (approx. $\pm 2$ p.p.), which is an indication that their accuracy is not affected by the increase of instances.

\begin{figure}[t]
    \centering
    \includegraphics[width=\columnwidth]{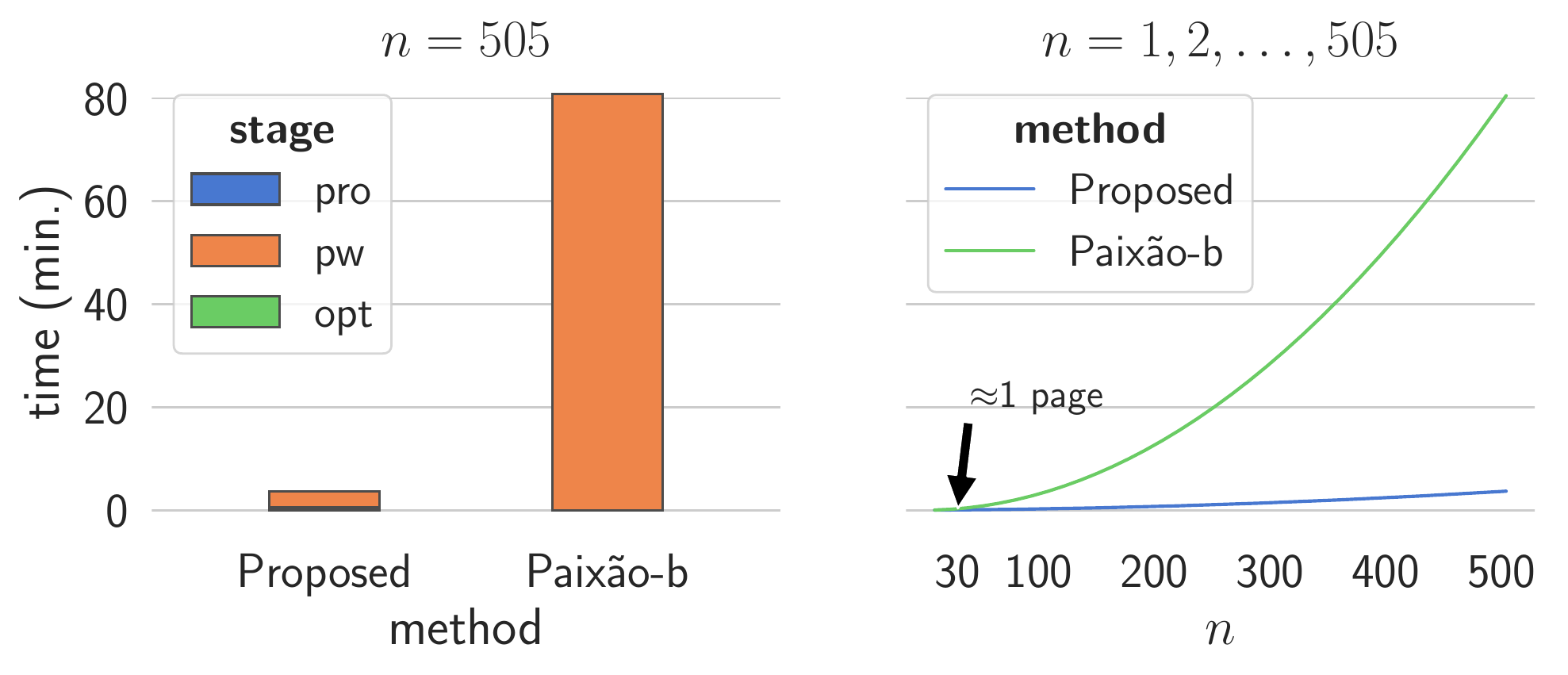} 
    \caption{Time performance for multi-page reconstruction. Left: the time demanded in each
        stage to reconstruct D2 entirely ($n=505$ shreds). Right: predicted processing time
        in function of the number of shreds.}
    \label{fig:exp2_time}
\end{figure}

\begin{figure}[t]
    \centering
    \includegraphics[width=\columnwidth]{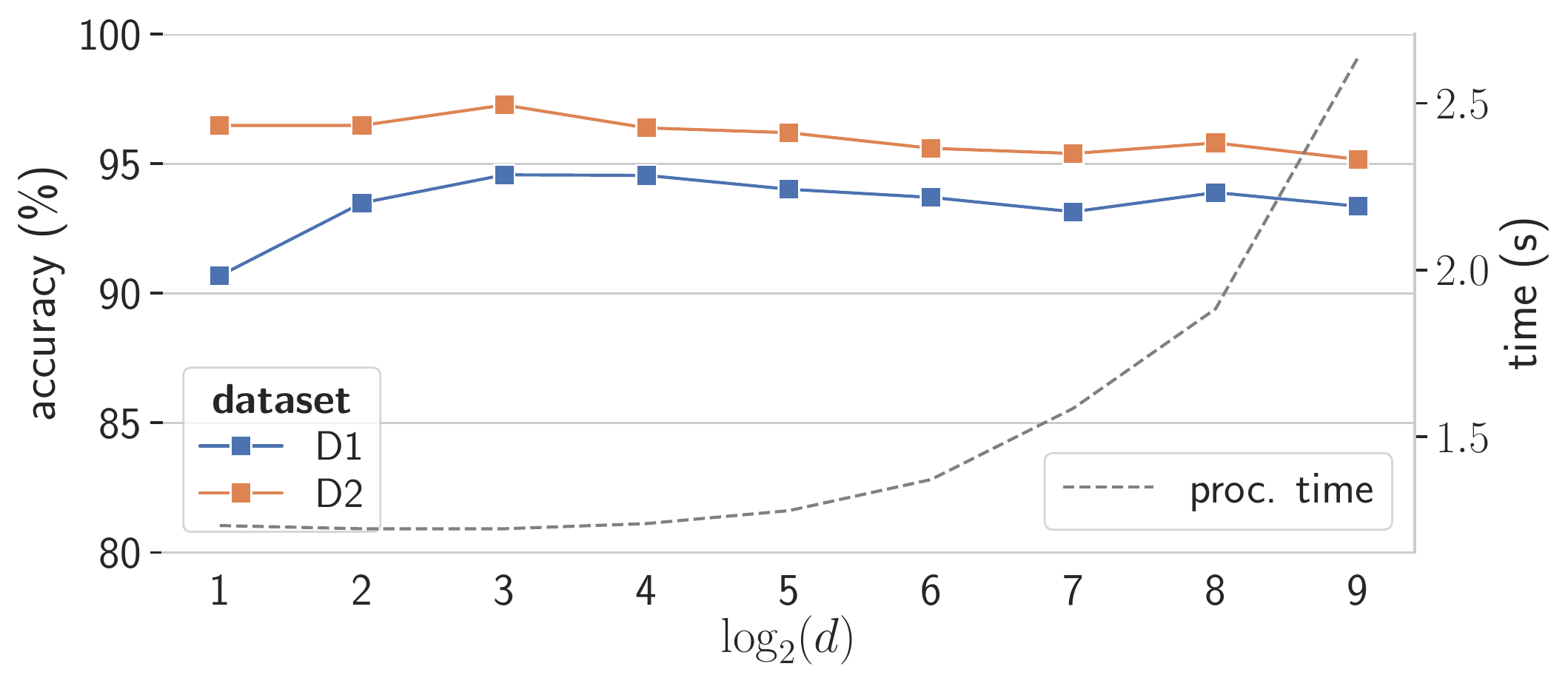} 
    \caption{Sensitivity analysis with respect to embeddings dimension ($d$). The best accuracy was
        observed for $d=8$: $94.57\%$ and $97.27\%$ for D1 and D2, respectively. This reduced embedded size
    yielded a reduction of $23\%$ on processing time.}
    \label{fig:ablation_feat_dim}
\end{figure}

Concerning time efficiency, however, the methods behave notably different, as evidenced in Figure~\ref{fig:exp2_time}. The left chart shows the average elapsed time of each stage to process the $505$ shreds of D2. In this context, with a larger number of shreds, the optimization cost became negligible when compared to the time required for pairwise evaluation. Remarkably, \base demanded more than $80$ minutes to complete evaluation, whereas our method took less than $4$ minutes (speed-up of approx. $22$ times). Based on the average time for the projection and the pairwise evaluation, estimation curves were plotted (right chart) indicating the predicted processing time in function of the number of shreds ($n$). Viewed comparatively, the growing of the proposed method's curve (in blue)
seems to be linear, although pairwise evaluation time (not the number inferences) grows quadratically with $n$. In summary, the greater the number of shreds, the higher the speed-up ratio.

\subsection{Sensitivity Analysis}

Figure~\ref{fig:ablation_feat_dim} shows, for single-page reconstruction, how accuracy and time processing (mean values over pages) are affected by the embedding dimension ($d$). Remarkably, projecting onto 2-D space ($d = 2$) is sufficient to achieve average accuracy superior to $90\%$. The highest accuracies were observed for $d=8$: $94.57\%$ and $97.27\%$ for D1 and D2, respectively. Also, the average reconstruction time for $d=8$ was $1.224$s, which represents a reduction of nearly $23\%$ when compared to the default value ($128$). For higher dimensions, accuracy tends to decay slowly (except for $d=256$). Overall, the results suggest that there is space for improvement on accuracy and processing time by focusing on small values of $d$, which will be better investigated in future work.

%% file: secs/6_conclusion.tex
\section{Conclusion}
\label{sec:conclusion}

This work addressed the problem of reconstructing mechanically-shredded text documents, more specifically the critical part of evaluating compatibility between shreds. Focus was given to the time performance of the evaluation. To improve it, we proposed a deep metric learning-based method as a compatibility function in which the number of inferences scales linearly rather than quadratically \cite{paixao2018deep} with the number of shreds of the reconstruction instance. In addition, the proposed method is trained with artificially generated data (i.e., does not require real-world data) in a self-supervised way (i.e., does not require annotation).

Comparative experiments for single-page reconstruction showed that the proposed method can achieve accuracy comparable to the state-of-the-art with a speed-up of $\approx 7$ times on compatibility evaluation. Moreover, the experimentation protocol was extended to a more realistic scenario in this work: multi-page multi-document reconstruction. In this scenario, the benefit of the proposed approach is even greater: our evaluation compatibility method takes less than 4 minutes for a set of 20 pages, compared to the approximate time of 1 hour and 20 minutes (80 minutes) of the current state-of-the-art (i.e., a speed-up of $\approx 22$ times), while preserving a high accuracy ($97.22\%$). Additionally, we show that the embedding dimension is not critical to the performance of our method, although a more careful tuning can lead to better accuracy and time performance.

%\textcolor{red}{I think we should try to move this out from the conclusion, maybe to the discussion. It feels bad to finish with this tone.} Despite the promising results, reconstruction of documents with much pictorial content should be deeper investigated. 

Future work should include the generalization of the proposed method to other types of cut (e.g., cross-cut and hand-torn), as well as to other problems related to jigsaw puzzle solving \cite{andalo2017}.

%% file: secs/acknowledgements.tex
\section*{Acknowledgements}

This  study  was  financed  in  part  by  the Coordena\c{c}\~{a}o de Aperfei\c{c}oamento de Pessoal de N\'{i}vel Superior - Brasil (CAPES) - Finance Code 001. We thank NVIDIA for providing the GPU used in this research. We also acknowledge the scholarships of Productivity on Research (grants 311120/2016-4 and 311504/2017-5) supported by Conselho Nacional de Desenvolvimento Cient\'{i}fico e Tecnol\'{o}gico (CNPq, Brazil).